\documentclass[manuscript, screen]{jair}
\setcopyright{cc}
\copyrightyear{2025}
\acmDOI{10.1613/jair.1.xxxxx}

\JAIRAE{Insert JAIR AE Name}
\JAIRTrack{} % Insert JAIR Track Name only if part of a special track

\makeatletter
\def\@mkfooters{}% disable the whole footer creation
\def\@mkjournalfoot{}%
\def\@acmArticle@printfooter{}%
\def\@acmArticle@footertext{}%
\makeatother

\RequirePackage[
  datamodel=acmdatamodel,
  style=acmnumeric,
  backend=biber,
  giveninits=true
  ]{biblatex}

\usepackage{subcaption} % subfigure
\usepackage{bm} % bold in equations
\usepackage{url} % handle URLs in bibliography
\usepackage{ltablex} % better space control for tables
\begin{document}

%\title{JAIR Example Template}
\title[Imitation Learning in the Deep Learning Era]{Imitation Learning in the Deep Learning Era: A Novel Taxonomy and Recent Advances}

%% used to denote shared contribution to the research and/or corresponding author.
\author{Iason Chrysomallis}
\authornote{Corresponding Author.}
\email{ichrysomallis@tuc.gr}
\affiliation{%
  \institution{Technical University of Crete}
  \city{Chania}
  \state{Crete}
  \country{Greece}
}

\author{Georgios Chalkiadakis}
\email{gchalkiadakis@tuc.gr}
\affiliation{%
  \institution{Technical University of Crete}
  \city{Chania}
  \state{Crete}
  \country{Greece}
}

%% The short list of authors must be made of the list of all authors' lastnames.
\renewcommand{\shortauthors}{Chrysomallis \& Chalkiadakis}
%% If this is too long and overlaps other information printed in the page headers, use
%\renewcommand{\shortauthors}{Xu et al.}

\begin{abstract}
Imitation learning (IL) enables agents to acquire skills by observing and replicating the behavior of one or multiple experts.
In recent years, advances in deep learning have significantly expanded the capabilities and scalability of imitation learning across a range of domains, where expert data can range from full state-action trajectories to partial observations or unlabeled sequences. 
Alongside this growth, novel approaches have emerged, with new methodologies being developed to address longstanding challenges such as generalization, covariate shift, and demonstration quality.
In this survey, we review the latest advances in imitation learning research, highlighting recent trends, methodological innovations, and practical applications. 
We propose a novel taxonomy that is distinct from existing categorizations to better reflect the current state of the IL research %field 
stratum 
and its trends. 
Throughout the survey, we critically examine the strengths, limitations, and evaluation practices of representative works, and we outline key challenges and open directions for future research.
\end{abstract}

%\begin{abstract}
%      A clear and well-documented \LaTeX\ document is presented as an
%  article formatted for publication by ACM in a conference proceedings
%  or journal publication. Based on the ``acmart'' document class, this
%  article presents and explains many of the common variations, as well
%  as many of the formatting elements an author may use in the
%  preparation of the documentation of their work.
%\end{abstract}

%% JAIR Note: 
%% Do not include ACM CCS Concepts or Keywords

%% To be updated by authors.
% \received{07 August 2025}
% \received[accepted]{TBD}

\maketitle

\section{Introduction}

{\em Imitation} has been extensively studied in psychology, where it is defined as the act of replicating a behavior in response to observing a similar action performed by another individual or animal~\cite{britannica}.
Imitation practices are widely observed throughout history and nature and play a vital role in survival, whether through biological evolution or social learning. 
It is a broad concept that encompasses a wide range of behaviors.

\begin{figure}[t]
  \centering
\includegraphics[width=0.7\textwidth]{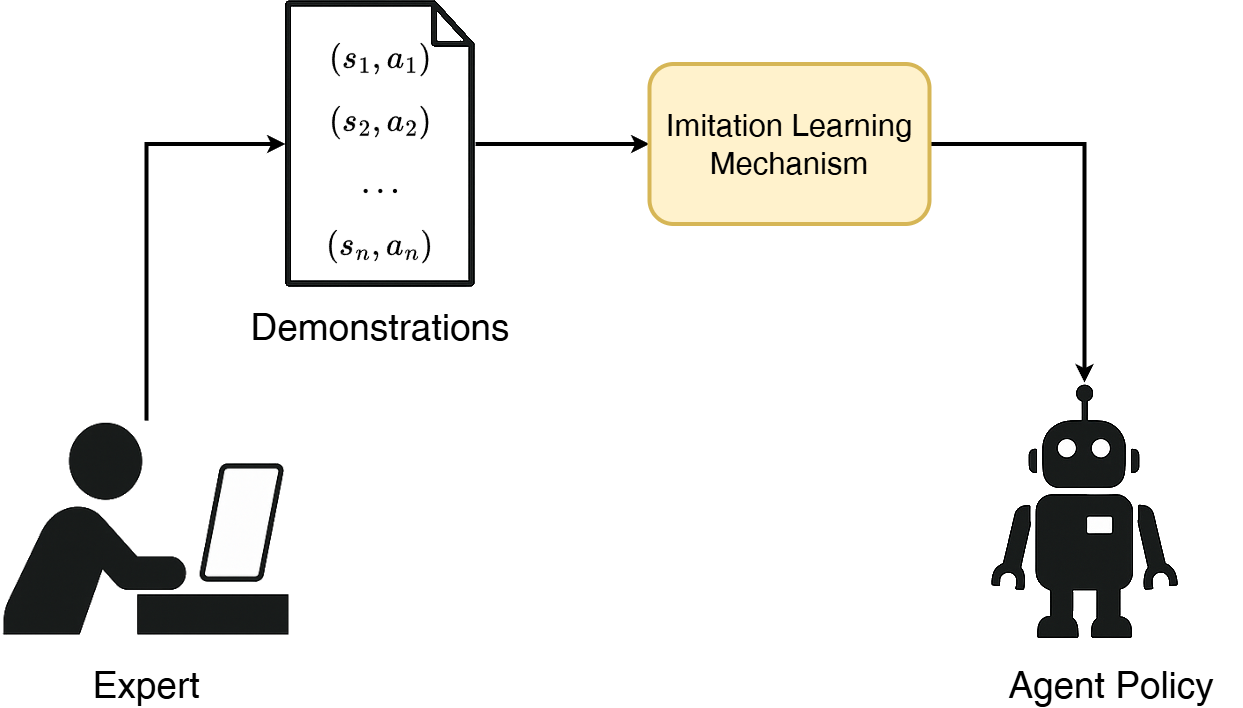}
\caption{Overview of imitation learning. An expert provides demonstrations in the form of trajectories, which are used by an imitation learning algorithm to train a policy that replicates the expert's behavior.}
\Description{A flowchart diagram showing the imitation learning process with four main components: an expert figure on the left providing demonstrations, an imitation learning mechanism that is designed to exploit those demonstrations, and an agent/robot on the right representing the trained policy, connected by arrows indicating the flow from expert through the algorithm to the final policy.}
\label{fig:imit_high}
\end{figure}

In the context of artificial intelligence and machine learning, imitation was initially defined as a set of techniques aimed at mimicking human behavior in specific tasks~\cite{hussein2017imitation}. 
In this setting, an artificial 
agent
is trained to perform a task by learning a mapping from observations to actions,
provided by humans.
Over time, learning from imitation or {\em imitation learning} has evolved beyond learning exclusively from humans. 
It is nowadays regarded more generally as a method for acquiring and developing new skills via the observation of the behavior of (or, more generally, from behavior-related data provided by) another agent possessing those skills~\cite{Billard2012}.
This gives rise to the following {\em imitation learning} definition, also visualized in Figure~\ref{fig:imit_high}:

\begin{definition}%[Imitation Learning]
\label{def}
\textbf{Imitation Learning (IL)} is a learning paradigm in which an agent seeks to acquire a policy by observing and imitating the behavior of 1...N proficient agents, collectively referred to as the \textbf{expert}. 
The expert provides a dataset $D$ of {\em expert-related behavioral data}, which may consist of, {\em wlog}, $M \in \mathbb{N}^+$ state-action pairs $D=\{(s_i,a_i)\}^M_{i=1}$; state-action-state transition tuples $D=\{(s_i,a_i; s^{\prime}_i)\}^M_{i=1}$; state-state transitions $D=\{(s_i,s'_i)\}^M_{i=1}$; or other supervisory signals.
Assuming an interactive environment,
the \textbf{learning agent} or \textbf{agent} is tasked with
employing this dataset 
in order to approximate the expert's behavioral policy $\pi_e: S \rightarrow \Delta(A)$, where $S$ is the environment's state space and $A$ the expert's action space.
\end{definition}

In this paper, we adopt the aforementioned definition.
Accordingly, there exists an experienced agent, human or otherwise, which we refer to as the \textit{expert}, that 
offers 
guidance through
the 
behavior-related data it provides. 
The learning 
\textit{agent}, observes 
this behavior-related data,
 and learns to perform the task by imitating the expert.

We note that there exist two other commonly used terms that refer to the same general framework (Table~\ref{tab:terms}): \textit{Learning from Demonstration} (LfD)~\cite{schaal1996learning,Mandlekar-RSS-20,argall2009survey,ravichandar2020recent,ijcai2019p882}; and \textit{Apprenticeship Learning}~\cite{apprent_inverse,abbeel2005exploration,syed2008apprenticeship}. 
However, in this paper, we use the widely adopted term {\em imitation learning} throughout.
\begin{tabularx}{\textwidth} { 
   >{\centering\arraybackslash}X 
   >{\centering\arraybackslash}X }
   \caption{Related terms for \textit{Imitation Learning}}\\
   \label{tab:terms}\\
    % \multicolumn{3}{|c|}{Country List} \\
    \hline
    Term & Notable works under this term \\
    \hline
    \hline
    Imitation Learning & ~\cite{zheng2022imitationlearningprogresstaxonomies,hussein2017imitation,gavenski2024imitation,schaal1999imitation,gail} \\
    \hline
    Learning from Demonstration & ~\cite{schaal1996learning,Mandlekar-RSS-20,argall2009survey,ravichandar2020recent,ijcai2019p882} \\
    \hline
    Apprenticeship Learning & ~\cite{apprent_inverse,abbeel2005exploration,syed2008apprenticeship} \\
    \hline
\end{tabularx}

Although the definition of imitation learning appears straightforward in principle, it can quickly become vague and ambiguous once we start considering more specific practical settings. 
Different assumptions about what the expert can provide lead to different formulations of how the agent should be trained. 
The guidance offered by the expert may take many forms, ranging from detailed first-person demonstrations with complete state-action information, to noisy third-person video observations, or even to verbal instructions. 
Each of these forms requires a different learning strategy and implies different challenges for interpretation and policy learning.
Another source of ambiguity lies in how closely the agent is expected to match the expert. 
In some cases, perfect imitation is required, while in others, a looser alignment in behavior or task outcome may be sufficient. 
This distinction has practical implications for algorithm design, as it determines the objective being optimized and the acceptable level of deviation from expert behavior. 
In certain tasks, replicating the precise trajectory may be important;  while in others, achieving a similar end result through a different path may be entirely valid.
It is also expected that some approaches will pursue goals that extend beyond the immediate scope of imitation. 
These may include optimizing performance beyond the expert's capabilities, adapting to constraints of the deployment environment, or integrating objectives such as safety, efficiency, or interpretability. 
These additions often arise from the nature of the domain or the generality of the problem being addressed.
As a result of all these factors---and depending on the assumptions made about the expert, the type of guidance available, and the goals of learning---the field of imitation learning includes a wide range of algorithms and methodological directions.

\paragraph{``Just mimicking an expert?'': A Common point of confusion} The goal of imitation learning, although seemingly clear by definition, often raises an important question. 
A common point of confusion is how an agent trained merely to mimic an expert can ever match or even outperform that expert. 
This is a fundamental concept within imitation learning, and, by design, the agent is not 
in general
intended to outperform the expert in terms of solution quality.
The main objective of an imitation learning model is to replicate the expert’s behavior, not to discover new or superior strategies.
On this matter of discussion, however, there are exceptions included in this survey that extend beyond simple behavioral mimicry ~\cite{chrysomallis2023deep,Chrysomallis_Chalkiadakis_Papamichail_Papageorgiou_2025,sasaki2021behavioral}.
These approaches do not settle for reproducing ``expert'' behavior in suboptimal cases, and instead attempt to refine or optimize the learned policy through additional mechanisms.
Nonetheless, even
when, as usual, its focus is not in surpassing the expert performance,
imitation learning offers several compelling advantages 
that make it a valuable and practical approach. 

One obvious benefit is the ability to transfer a human policy to an autonomous agent, thereby eliminating the need for a human to be available at decision time. 
Beyond that, a more substantial advantage, particularly in the absence of human experts, is the {\em generalization} of the expert policy.
In such cases, the expert is typically an algorithm, not a neural network, and is often tailored to specific instances or tightly coupled to particular constraints and assumptions encoded in its design. 
An imitation learning model, by contrast, learns a general mapping from states to actions, enabling it to scale to problem instances far larger than those the expert could practically handle. 
Moreover, by learning the patterns in how the expert behaves, the model becomes better at responding to unseen situations. 
These cases may not have been covered by the expert, but they still share structural similarities, allowing the model to generalize and handle them more reliably.

Another notable benefit is efficiency. 
Algorithmic experts are frequently slow to execute, memory-intensive, and difficult to integrate into real-time systems or resource-constrained environments. 
An imitation learning policy, once trained, can be deployed as a compact neural network with fixed memory and prediction time, regardless of the input size.
Training efficiency is also typically higher for imitation learning models, which are faster and less computationally expensive compared to alternatives such as \textit{reinforcement learning} (RL)~\cite{sutton1998reinforcement}. 
RL depends on prolonged interaction with the environment and iterative updates based on collected rewards, often over thousands or millions of steps.
In imitation learning, the training process follows a clear, supervised direction using expert demonstrations, which leads to much faster and more focused learning.

\paragraph{Do I call you an ``expert'' or a ``teacher''?}

In this paper we avoid using the terms ``teacher'' and ``student'' due to their strong association with \textit{knowledge distillation}~\cite{hinton2015distilling}, which should not be confused with imitation learning. 
In the context of knowledge distillation, the teacher is typically a neural network, and the student is another network trained to replicate the teacher’s output. 
This replication goes beyond the teacher's final decision, and includes the full distribution of its confidence across all possible actions. 
The objective is to approximate this entire probability distribution, rather than to reproduce specific actions.

While this distinction may not be immediately apparent on an intuitive level, it results in significant differences in how the agent is trained. 
In imitation learning, the agent receives only the selected action taken by the expert
% \footnote{We note that when an ``expert'' does not necessarily follow an optimal policy, a perhaps more appropriate term for it is ``mentor''~\cite{chrysomallis2023deep,Chrysomallis_Chalkiadakis_Papamichail_Papageorgiou_2025}.} 
in a given state and learns to mimic that specific action. 
By contrast, knowledge distillation provides the student with a complete distribution over possible actions, and the student learns to match that distribution. 

For illustration, consider a simplified scenario in which a vehicle approaches a crossroad with the option to turn left or right. 
An imitation learning algorithm would receive the expert's choice {\em turn right} as a ``hard label''. 
In a knowledge distillation setting, the model would instead receive a ``soft label'', such as an ``$80\%$ probability of turning right''  and ``$20\%$ probability of turning right'', and aim to match that distribution.
In this paper, we focus primarily on imitation learning algorithms, with the exception of~\cite{ijcai2021p457}, which is included due to its relevance to the specific topic (global consistency on sequential recommendation tasks) of the corresponding section.
Thus, and for the sake of consistency, we adopt the term expert throughout this survey.

\begin{figure}[t]
  \centering
\includegraphics[width=0.7\textwidth]{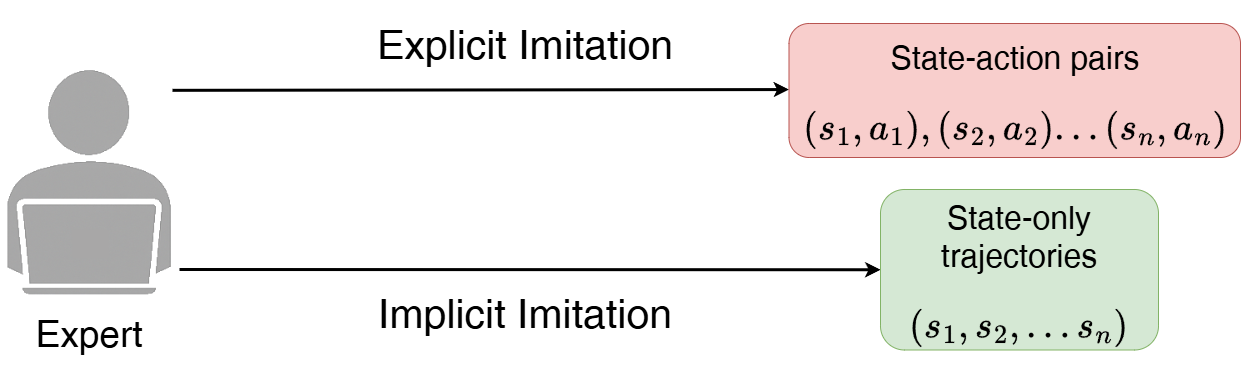}
\caption{Differences between explicit and implicit imitation. In both cases, the expert provides a dataset, but in explicit imitation this contains state-action pairs, whereas in implicit imitation only state transitions are available.}
\Description{A diagram showing an expert on the left with two arrows pointing right. The top arrow is labeled ``Explicit Imitation'' and points to a box containing state-action pairs. The bottom arrow is labeled ``Implicit Imitation'' and points to a box containing state-only trajectories.}
\label{fig:expl_impl}
\end{figure}

\subsection{Explicit Imitation vs Implicit Imitation vs Inverse Reinforcement learning} 
\label{subsec:EIvsIIvsIR}

As mentioned earlier, even though the definition and goals of imitation learning are straightforward, there are important factors that influence how the problem is approached. 
One of the most fundamental aspects is the form of provided expert data, which can vary significantly. 
To make these distinctions clear, we introduce a few terms that will be used throughout this paper (Figure~\ref{fig:expl_impl}).
We refer to expert data that includes both the visited states and the corresponding actions taken---typically represented as $(s, a)$ pairs, or in some cases $(s, a, s')$ including the 
%next 
state $s'$
resulting from the expert executing $a$ at state $s$---as \textit{demonstrations}. 
This form of expert guidance reflects the original and most intuitive approach to imitation learning, where both what the expert %saw
encountered
and what they did are made available. 
We classify IL research employing this type of expert guidance as \textit{explicit imitation}, where the agent is trained using full access to the expert’s decision-making process.
By contrast, the expert 
might
(be able or willing to)
provide only a sequence of its
observed states or state transitions $(s,s')$, referred to as \textit{observations}, 
without revealing the actions that caused them.
IL research making use of this type of data, falls under what we term 
\textit{implicit imitation}~\cite{price2003accelerating,price1999implicit,price2001imitation,chrysomallis2023deep,Chrysomallis_Chalkiadakis_Papamichail_Papageorgiou_2025}, also known as 
\textit{Learning from Observation} (LfO)~\cite{NEURIPS2021_868b7df9,ijcai2018p687,torabi2018generative,liu2024imitation,ijcai2019p882,household,GONZALEZ2022117167}.
% ~\cite{GONZALEZ2022117167,NEURIPS2021_868b7df9,ijcai2018p687,torabi2018generative,liu2024imitation,ijcai2019p882, household}. 
In this setting, the learning agent must infer how to act based solely on the way the expert moved through the environment, without knowing which actions were taken at each step.

Another important factor within the scope of imitation learning is the objective of learning itself. 
While the most intuitive approach focuses on answering \textit{``how''} an expert solves a task and learning to reproduce the expert's behavior, an alternative perspective asks \textit{``why''} the expert behaves in a certain way. 
This shifts the focus from behavior replication to understanding the motivations behind the behavior.
This approach offers a deeper level of interpretability, and facilitates better generalization---especially in settings where direct imitation may fail due to changes in environment dynamics, or task specifications~\cite{apprent_inverse}.
This domain is studied under the framework of \textit{Inverse Reinforcement Learning}~\cite{ng2000algorithms}, which aims to uncover the underlying reward function that explains the expert’s behavior.
Rather than directly learning and replicating the expert policy, it attempts to infer the objective that the expert is optimizing. 
Once this reward function is recovered, standard reinforcement learning algorithms can be applied to derive a policy that achieves similar (or even superior) performance, potentially under new or varied conditions~\cite{finn2016connection}.
This approach to imitation learning is particularly useful when the goal is to interpret, generalize, or transfer expert strategies, as well as foundational in interpretable AI efforts, where understanding the motivation behind decisions is essential~\cite{doshi2017towards}.

Lastly, an additional consideration that applies across all forms of imitation, whether explicit imitation, implicit imitation, or inverse reinforcement learning, is how closely the learned policy should follow the expert’s behavior.
While the majority of imitation learning approaches aim to match or approximate expert performance, there is growing recognition that suboptimal experts may introduce bias or errors into the dataset. 
This can lead the learning agent to replicate behaviors that are not ideal, particularly in edge cases. 
To address this, some methods explicitly account for the possibility of suboptimal demonstrations by controlling or adjusting the expert’s influence during training~\cite{chrysomallis2023deep,Chrysomallis_Chalkiadakis_Papamichail_Papageorgiou_2025,sasaki2021behavioral,sun2022deterministic,zhou2024rethinking}. 
These techniques seek to balance imitation with correction, depending on how reliable the expert data is believed to be.

\subsection{Our Focus and a Novel Taxonomy}

Multiple surveys have been published in recent years on imitation learning, generally following a common pattern of introducing the foundational concepts across various types of imitation learning approaches and then narrowing the focus to a specific category. 
The works in~\cite{zheng2022imitationlearningprogresstaxonomies,hussein2017imitation} provide a comprehensive overview of the first wave of imitation learning methods and serve as a valuable reference point and benchmark for the more recent developments that define the current era. 
The survey in~\cite{ravichandar2020recent} concentrates on applications in robotics, addressing the unique challenges and constraints that arise specifically within that domain. 
The work in~\cite{ijcai2019p882} limits its scope to learning from observation techniques, instead of a more general discussion across the broader field of imitation learning.
More recently,~\cite{gavenski2024imitation} identifies ongoing trends in the area and, importantly, proposes the first taxonomy of imitation learning environments based on the role of the environment in the learning process.
Regarding transfer learning, a closely related field to imitation learning, ~\cite{da2019survey} offers an in-depth survey of multi-agent specific implementations and knowledge sharing strategies.
In this paper, we shift the focus towards recent advances in imitation learning, as appearing in peer-reviewed works from the last few years. 
We introduce a novel taxonomy, which diverges %slightly 
from those previously proposed~\cite{zheng2022imitationlearningprogresstaxonomies,hussein2017imitation,ravichandar2020recent,gavenski2024imitation}, in order to better match the focus and direction of recent research.

\begin{figure}[!ht]
  \centering
\includegraphics[width=0.75\textwidth]{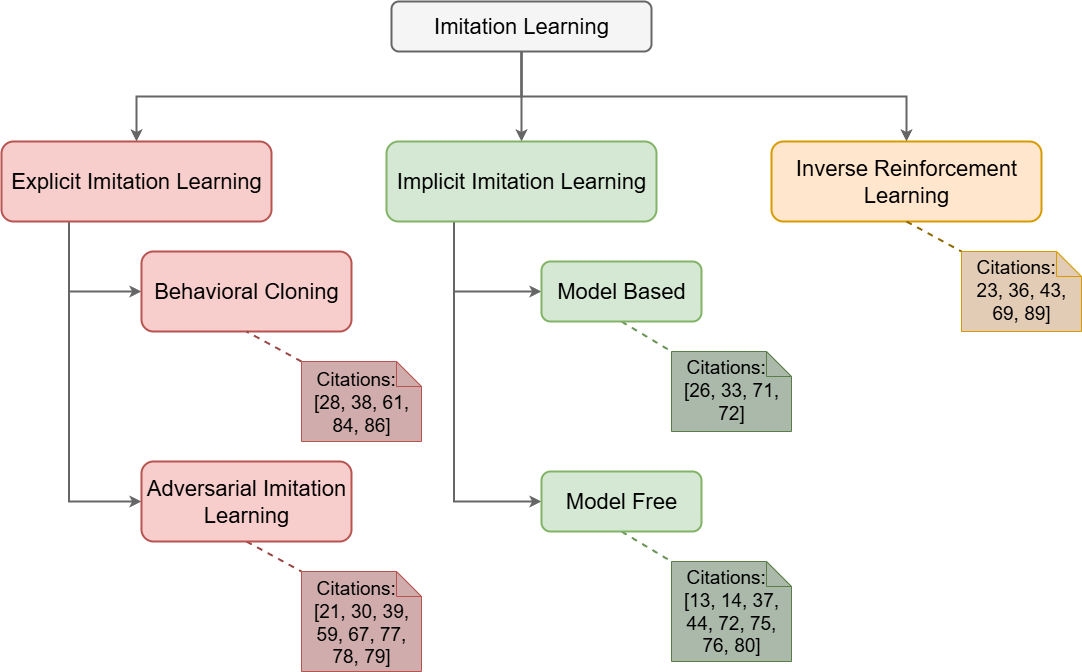}
\caption{An imitation learning taxonomy}
\Description{A hierarchical tree diagram showing our proposed taxonomy of imitation learning. At the top is \textit{Imitation Learning} which branches into three main categories: \textit{Explicit Imitation Learning}, \textit{Implicit Imitation Learning}, and \textit{Inverse Reinforcement Learning}. The explicit imitation branch splits into \textit{Behavioral Cloning} and \textit{Adversarial Imitation Learning}, while the implicit branch splits into \textit{Model Based} and \textit{Model Free}. Each subcategory has an associated citation box containing reference numbers.}
\label{fig:taxonomy}
\end{figure}

Indeed,
given that imitation learning is a rapidly evolving field, with shifting trends and %central
focal
points over time, it is only natural that new taxonomies emerge to better reflect current research directions. 
As such, in this work,
we propose our own taxonomy, 
whose structure
emphasizes recent research trends in the imitation learning literature.
One of the most significant trends in recent years is the rise of {\em implicit} imitation methods~\cite{ijcai2019p882,GONZALEZ2022117167,household,wang2024diffail,pmlr-v229-wang23a,NEURIPS2021_868b7df9,chrysomallis2023deep,Chrysomallis_Chalkiadakis_Papamichail_Papageorgiou_2025}. 
This motivates us to draw a primary distinction between {\em explicit} and {\em implicit} imitation learning approaches; while another persisting major IL category is the {\em inverse reinforcement learning} one.
Our complete IL taxonomy, which also gives rise to the structure of this survey paper, is depicted in Figure~\ref{fig:taxonomy}. 
We believe that this taxonomy does reflect recent developments in the IL literature; and anticipate that it provides a clear and intuitive structure for readers of this paper. 
The distinguishing aspects among these categories are discussed in Section~\ref{subsec:EIvsIIvsIR} above.
We now proceed to further describe 
our proposed taxonomy; we note that  the key papers in each category are reviewed in the corresponding sections of this paper.

In {\em explicit imitation}, the expert provides both state and action information. 
Within this setting, we focus on two major subcategories, \textit{Behavioral Cloning} and \textit{Adversarial Methods}. 
Behavioral Cloning represents the foundational approach in the deep learning era of imitation learning. 
A large body of work builds on Behavioral Cloning, either by extending its capabilities or by addressing its limitations. 
In parallel, adversarial methods, inspired by the seminal \textit{Generative Adversarial Imitation Learning} (GAIL) framework, have led to a wide array of imitation learning techniques that incorporate adversarial training. 
Many recent works propose variations of GAIL or introduce new adversarial formulations to overcome its known challenges.

In {\em implicit imitation}, the agent observes only the expert’s state transitions, without access to their actions. 
This setting is less strictly categorized, so we divide the methods more broadly into model-based and model-free approaches. 
Model-based methods, particularly those relying on inverse dynamics models, were prominent in the earlier stages of implicit imitation research~\cite{ijcai2019p882}. 
While such methods have become less dominant in recent years, a few notable works continue to explore model-based strategies from new perspectives~\cite{household,GONZALEZ2022117167}. 
By contrast, model-free approaches, including both adversarial and reinforcement learning-based methods, have seen significant growth in the past few years. 
These methods avoid modeling dynamics explicitly and instead focus on learning policies or value functions directly from observations.

Finally, we include \textit{Inverse Reinforcement Learning} (IRL) as a distinct category. 
IRL approaches imitation by inferring the expert’s underlying reward function, using it as a proxy to understand and replicate their behavior. 
IRL was a groundbreaking shift in how imitation learning was conceptualized~\cite{zhifei2012survey,arora2021survey,adams2022survey}, and although interest in it fluctuated over time, recent works continue to explore its potential, often with new tools or perspectives that build on its fundamental strengths.

A full list of the papers that are thoroughly 
reviewed in this survey is provided in Table~\ref{tab:list}. 
Each entry is associated with its corresponding category and subcategory under the proposed taxonomy, the type of expert data used (such as state-action pairs or state-only transitions), and the nature of the agent’s learning setup, indicating whether training occurs through direct interaction with the environment or from fixed demonstrations.
We will be primarily reviewing post-2020 imitation learning papers; however, we also include principal representative works~\cite{bain1995framework,gail,ijcai2018p687,torabi2018generative,pmlr-v48-finn16} within each (sub-)category of our taxonomy, as these papers 
are essential to understanding the evolution of the field. 
Many foundational methods emerged during the early stages of the deep learning era, and we argue that understanding these earlier contributions is critical to appreciating the challenges that modern approaches aim to address.

% expl BC: ~\cite{hejna2023improving,pfss,sasaki2021behavioral,10.1609/aaai.v37i9.26305,ijcai2021p457}\\
% expl adv: ~\cite{ijcai2020p405,gail,li2017infogail,Ruan_Di_2022,sun2022deterministic,wang2021reinforced,Wang_Gao_Wu_Jin_Yao_Li_2023,wang2024exploring}\\
% impl based: ~\cite{GONZALEZ2022117167,household,ijcai2018p687,torabi2018generative}\\
% impl free: ~\cite{Chrysomallis_Chalkiadakis_Papamichail_Papageorgiou_2025,chrysomallis2023deep,NEURIPS2021_868b7df9,liu2024imitation,torabi2018generative,wang2024diffail,pmlr-v229-wang23a,wu2021textgail}\\
% IRL: ~\cite{pmlr-v48-finn16,kumar2023graph,NEURIPS2022_d842425e,swamy2023inverse,zhou2024rethinking}

\begin{tabularx}{\textwidth} { 
   >{\centering\arraybackslash}X 
   >{\centering\arraybackslash}X 
   >{\centering\arraybackslash}X 
   >{\centering\arraybackslash}X 
   >{\centering\arraybackslash}X 
   >{\centering\arraybackslash}X }
    % \multicolumn{3}{|c|}{Country List} \\
    \caption{Full list of papers covered in this survey, with their (sub-)category, expert data type, and (online/offline) training setup. {\em BC} stands for {\em Behavioral Cloning}.}\label{tab:list}\\
    \hline
    Citation & Category & Subcategory & Expert Data Type & Online/Offline & Year \\
    \hline
    \hline
    ~\cite{bain1995framework} & Explicit & BC & Demonstration & offline & 1995 \\
    \hline
    ~\cite{ijcai2021p457} & Explicit & BC & Demonstration & offline & 2021 \\
    \hline
    ~\cite{sasaki2021behavioral} & Explicit & BC & Demonstration & offline & 2021 \\ 
    \hline
    ~\cite{hejna2023improving} & Explicit & BC & Demonstration & offline & 2023 \\ 
    \hline
    ~\cite{10.1609/aaai.v37i9.26305} & Explicit & BC & Demonstration & offline & 2023 \\ 
    \hline
    ~\cite{pfss} & Explicit & BC & Demonstration & offline & 2024 \\
    \hline
    ~\cite{yu2024usn} & Explicit & BC & Demonstration & - & 2024 \\
    \hline
    ~\cite{gail} & Explicit & Adversarial & Demonstration & online & 2016 \\ 
    \hline
    ~\cite{li2017infogail} & Explicit & Adversarial & Demonstration & online & 2017 \\ 
    \hline
    ~\cite{ijcai2020p405} & Explicit & Adversarial & Demonstration & online & 2020 \\
    \hline
    ~\cite{wang2021reinforced} & Explicit & Adversarial & Demonstration & offline & 2021 \\ 
    \hline
    ~\cite{Ruan_Di_2022} & Explicit & Adversarial & Demonstration & online & 2022 \\ 
    \hline
    ~\cite{sun2022deterministic} & Explicit & - & Demonstration & online & 2022 \\  
    \hline
    ~\cite{Wang_Gao_Wu_Jin_Yao_Li_2023} & Explicit & Adversarial & Demonstration & online & 2023 \\
    \hline
    ~\cite{wang2024exploring} & Explicit & Adversarial & Demonstration & online & 2024 \\ 
    \hline
    ~\cite{ijcai2018p687} & Implicit & Model-Based & Observation & offline & 2018 \\ 
    \hline
    ~\cite{GONZALEZ2022117167} & Implicit & Model-Based & Observation & offline & 2022 \\ 
    \hline
    ~\cite{household} & Implicit & Model-Based & Observation & offline & 2024 \\
    \hline
    ~\cite{torabi2018generative} & Implicit & Model-Free & Observation & online & 2018 \\
    \hline
    ~\cite{NEURIPS2021_868b7df9} & Implicit & Model-Free & Observation & online & 2021 \\ 
    \hline
    ~\cite{chrysomallis2023deep} & Implicit & Model-Free & Observation & online & 2023 \\ 
    \hline
    ~\cite{pmlr-v229-wang23a} & Implicit & Model-Free & Video & offline & 2023 \\ 
    \hline
    ~\cite{liu2024imitation} & Implicit & Model-Free & Observation & online & 2024 \\   
    \hline
    ~\cite{wang2024diffail} & Implicit & Model-Free & Observation & online & 2024 \\ 
    \hline
    ~\cite{wu2021textgail} & Implicit & Model-Free & Text & - & 2024 \\ 
    \hline
    ~\cite{Chrysomallis_Chalkiadakis_Papamichail_Papageorgiou_2025} & Implicit & Model-Free & Observation & online & 2025 \\ 
    
    \hline
    ~\cite{pmlr-v48-finn16} & IRL & - & Demonstration & online & 2016 \\ 
    \hline
    ~\cite{NEURIPS2022_d842425e} & IRL & - & Demonstration & offline & 2022 \\ 
    \hline
    ~\cite{swamy2023inverse} & IRL & - & Demonstration & offline & 2023 \\ 
    \hline
    ~\cite{kumar2023graph} & IRL & - & Video & offline & 2023 \\ 
    \hline
    ~\cite{zhou2024rethinking} & IRL & - & Observation & offline & 2024 \\ 
    \hline
\end{tabularx}

\subsection{Paper structure}

Reflecting the evolution of the field, modern research trends prioritize challenges tied to real-world applicability, including learning from limited or imperfect demonstrations and improving stability in offline or constrained environments.
This survey offers a deep look at how these challenges are being approached currently in the literature.
Each reviewed approach is presented in appropriate detail, with attention to the underlying motivation as well as to the specific training methodologies employed.

The remainder of this survey is organized as follows.
We begin with Section~\ref{section:preliminaries}, which presents core preliminaries. This includes essential background definitions, a historical timeline of the field, and an overview of real-world applications where imitation learning has been employed.
In the subsequent sections, we review representative papers based on our proposed taxonomy.
Section~\ref{section:explicit} introduces explicit imitation, covering both foundational behavioral cloning approaches and adversarial methods.
In Section~\ref{section:implicit}, we turn to implicit imitation, distinguishing between model-based and model-free variants depending on how expert behavior is inferred from limited information.
Section~\ref{section:irl} completes the main review by examining works in inverse reinforcement learning, where the focus shifts from replicating behavior to uncovering the reward or constraint structure behind it.
Afterwards, in Section~\ref{section:challenges} we outline some challenges present in imitation learning, and we point to areas that what we deem worthy exploring based on their relevance and current research endeavors.
Finally, in Section~\ref{section:conclusion}, we summarize our key takeaways and highlight the contributions of this survey.

\section{Preliminaries}
\label{section:preliminaries}
Having introduced the core concepts of imitation learning and outlined the focus of this survey, it is important to first establish a few preliminaries before reviewing recent works. In particular, we provide the definition of a Markov Decision Process (MDP), as it provides the formal framework within which most imitation learning methods are developed. Following that, we also provide a brief overview of {\em deep reinforcement learning}.
This is necessary so that
the formulation of, and the optimization process employed by several algorithms in this survey, becomes clear to the reader.
Following that, we provide a brief overview of imitation learning application domains.
We conclude this section with a timeline of imitation learning research.

\subsection{Markov Decision Processes} 
A \textit{Markov Decision Process (MDP)}~\cite{bellman1957markovian,putermanMDPs} is a mathematical framework commonly used to represent not only imitation learning problems but a wide range of decision-making / decision-theoretic planning tasks. 
At its core, an MDP describes how an agent interacts with an environment over time by moving through different states, taking actions, and receiving feedback.
An MDP is typically defined by the tuple $\langle S, A, T, R, \gamma\rangle$.
$S$ denotes the state space representing all possible configurations of the environment or situations that the agent may encounter. 
To influence the environment, the agent has an action space $A$, which includes all possible actions the agent can take. 
The behavior of the environment in response to these actions is captured by the transition model $T$, which defines the probability or rule that determines the next state, given the current state and chosen action.
To guide learning, the framework includes a reward function $R$, which provides a numerical signal indicating the desirability of a state-action transition, typically in the form of positive or negative feedback.
Note however that, in imitation learning, the reward signal is often either unavailable or unused, as the focus shifts from maximizing rewards to replicating expert behavior. 
The last element in the MDP tuple, $\gamma$ denotes a {\em discount factor} that is  bounded within $[0,1)$, especially useful in the context of infinite horizon problems. This factor
determines the importance of future rewards relative to immediate ones. 
Intuitively, a low $\gamma$ value causes the agent to prioritize short-term outcomes, while a higher value places more weight on the inclusion of long-term rewards as well.

With an MDP formulation at hand, one is able to employ an (e.g., dynamic programming) algorithm of choice in order to compute an optimal {\em policy} 
which defines the behavior of the agent. 
A \textit{policy} is a function that maps states to actions, determining what the agent should do at a given state. 
In deterministic settings, this function is expressed as $\pi(s)=a$, meaning the agent always takes a specific action in a given state. 
In stochastic settings, the policy defines a probability distribution over the action space for each state, representing the likelihood of taking each action. 
The policy captures the decision-making strategy of the agent, and training a policy to perform well is the central goal of most decision-making algorithms. 
In imitation learning, this goal is recast as learning a policy that closely replicates the behavior of an expert.
A \emph{value function}~\cite{sutton1998reinforcement} is often used in this context to estimate how good it is for the agent to be in a particular state (or to take a specific action in that state), typically in terms of expected future rewards.
Many decision-making algorithms leverage value functions to guide policy improvement and evaluate performance.

\subsection{Deep Reinforcement Learning}

Now, when $R$ or $T$ in an MDP are unknown, we are in {\em reinforcement learning (RL)}~\cite{sutton1998reinforcement} territory, where the aim is essentially to perform decision-theoretic planning (i.e., to determine an optimal policy) by trial-and-error in an MDP environment to whose transition or reward dynamics we have no access to.
{\em Deep reinforcement learning (DRL)} enhances RL algorithms' learning and generalization abilities via making use of modern (deep) neural networks that can efficiently approximate the value and/or policy functions.

DRL is ostensibly a fundamentally different paradigm from imitation learning, aiming to learn optimal policies through interaction with the environment and reward maximization rather than expert supervision.
Regardless, it plays a key role in many modern imitation learning approaches. 
Several algorithms adopt DRL-style training, using expert demonstrations to initialize or guide policy learning while updating the policy through standard DRL mechanisms. 
In some cases, these methods generate their own synthetic trajectories (states and actions) while using reward functions derived from the expert data or other heuristics. 
Additionally, some imitation learning techniques explicitly integrate DRL as part of a hybrid approach, combining imitation learning techniques with DRL practices. 
To help understand these methods, we briefly introduce the core principles of widely used DRL algorithms that commonly appear in imitation learning research.

The \textit{Deep Q-Network}~\cite{DQN} (DQN) algorithm extends Q-learning to high-dimensional state spaces with discrete action spaces using deep neural networks as function approximators. 
The agent learns a Q-function $Q(s,a;\theta)$, a neural network with parameters $\theta$ which estimates the expected cumulative reward of taking action $a$ in state $s$, following the optimal policy thereafter. 
DQN minimizes the temporal difference (TD) loss:
\begin{equation}
    L(\theta)=E_{(s,a,r,s')} D[(Q(s,a;\theta) - y)^2]
\end{equation}
where the target $y=r+\gamma max_{a'} Q(s',a';\theta^-)$, and $\theta^-$ are the parameters of a target network updated less frequently than $\theta$. 
DQN is commonly used in discrete action spaces and serves as a backbone for various implicit imitation methods where expert transitions are used to guide or initialize Q-values.

\textit{Deep Deterministic Policy Gradient}~\cite{lillicrap2015continuous} (DDPG) is an actor-critic algorithm for continuous action spaces. 
It maintains an {\em actor network} $\pi(s;\theta)$ that
determines a policy and outputs corresponding actions; and a {\em critic network} $Q(s,a;\theta^Q)$ that evaluates action values. 
The critic is updated using a TD loss similar to DQN:
\begin{equation}
    L(\theta^Q)=E_{(s,a,r,s')}\left[Q(s,a;\theta^Q)-\left(r+\gamma Q(s',\pi(s';\theta);\theta^Q)\right)\right]^2
\end{equation}
The actor is updated via the deterministic policy gradient:
\begin{equation}
    \nabla_{\theta} J \approx E_{s} \left[ \nabla_a Q(s, a; \theta^Q) \big|_{a = \pi(s;\theta)} \nabla_{\theta} \pi(s;\theta) \right]
\end{equation}

\subsection{Imitation Learning in Practice} 

Imitation learning has been widely explored across a range of domains, each leveraging expert behavior to train autonomous agents. 
Perhaps the most intuitive application is robotics, where early work focused on transferring human skills to robotic systems capable of mimicking demonstrated movements~\cite{argall2009survey,ravichandar2020recent,schaal1999imitation,mandlekar2020iris,household,Billard2012}
This foundational setting highlighted the potential of imitation learning for motor control and low-level policy transfer.
Autonomous driving represents another early and impactful domain, beginning with systems like ALVINN~\cite{pomerleau1988alvinn} that directly mapped visual input to control commands. 
This direction has since evolved into a rich field of research~\cite{chrysomallis2023deep,Ruan_Di_2022,chib2023recent}, refining mimicking procedures and addressing the safety and generalization challenges of real-world driving.
In recent years, video games and physics-based simulation environments have become dominant testbeds for imitation learning, due to their flexibility, reproducibility, and compatibility with deep reinforcement learning architectures~\cite{gavenski2024imitation,gail,torabi2018generative,Chrysomallis_Chalkiadakis_Papamichail_Papageorgiou_2025}. 
These domains allow agents to learn complex behaviors in safe, controlled settings, and serve as important benchmarks for evaluating generalization and robustness.
Imitation learning has also been applied to healthcare~\cite{GONZALEZ2022117167,yu2019inverse}, where expert demonstrations, such as treatment decisions, can guide agents toward medically sound policies.
These applications aim to reduce human error, enhance consistency, and provide data-driven guidance by modeling sequences of expert actions that inform treatment strategies.
Finally, in the latest trend of natural language processing, imitation learning techniques have been applied to text generation tasks~\cite{li2019dialogue,wu2021textgail}, including dialogue modeling and story generation. 
These methods learn to produce coherent text by imitating expert-written examples.

\subsection{Imitation Learning Timeline}
\begin{figure}[t!]
  \centering
\includegraphics[width=1\textwidth]{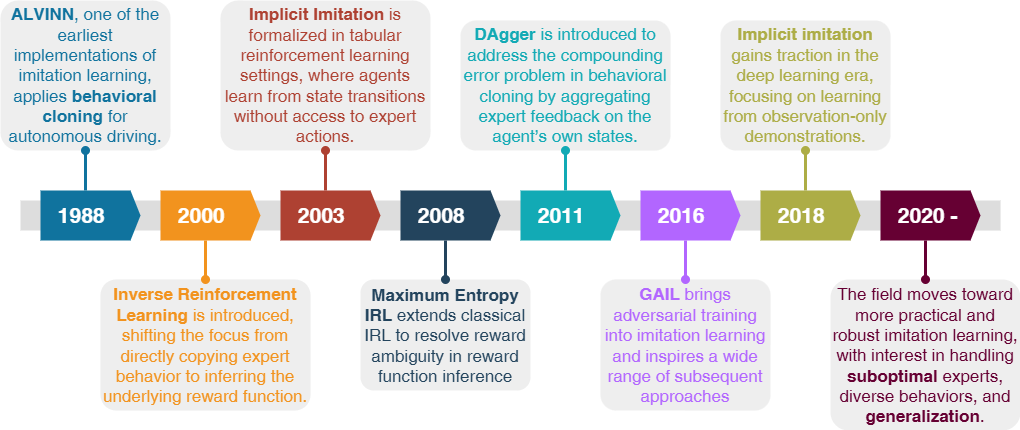}
\caption{Historical timeline of imitation learning}
\Description{A horizontal timeline showing the evolution of imitation learning from $1988$ to $2020+$. The timeline consists of connected boxes, each representing a year ($1988$, $2000$, $2003$, $2008$, $2011$, $2016$, $2018$,$ 2020-$). Above and below each year are boxes describing key developments: ALVINN and behavioral cloning (1988), Inverse Reinforcement Learning introduction (2000), Implicit Imitation formalization (2003), Maximum Entropy IRL (2008), DAgger for addressing compounding errors (2011), GAIL bringing adversarial training (2016), focus on observation-only learning (2018), and recent developments in practical applications (2020+).}
\label{fig:history}
\end{figure}

Imitation learning has a long and evolving history, originating in early robotics and control systems and gradually expanding into a rich and diverse field that now plays a central role in modern machine learning (Figure~\ref{fig:history}). 
From the early foundations of \textit{Behavioral Cloning}~\cite{pomerleau1988alvinn} in the late 80s, which focused on directly mapping states to actions, to the introduction of \textit{Inverse Reinforcement Learning}~\cite{ng2000algorithms} in 2000, which sought to infer the expert’s underlying reward function rather than simply mimic their behavior, the field 
steadily evolved in both scope and depth. 
The early $2000$s saw the emergence of the first works on \textit{implicit imitation}~\cite{price2003accelerating, price1999implicit, price2001imitation},
which 
introduced 
the idea of learning 
{\em only} 
from state observations
in (tabular) reinforcement learning settings. 
Subsequent advances refined both major paradigms, \textit{Max Entropy IRL}~\cite{ziebart2008maximum} improved reward function inference by addressing ambiguity in expert behavior, while \textit{DAgger}~\cite{ross2011reduction} tackled the compounding error issue in \textit{Behavioral Cloning} by incorporating expert feedback on agent-visited states. 
The deep learning era brought a major turning point with the introduction of adversarial learning into imitation~\cite{gail}, which inspired a broad family of imitation learning methods that remain highly influential today. 
Around the same time, research into implicit imitation began to gain momentum~\cite{ijcai2019p882}. 
Today, with foundational methods well established, the field is moving toward more practical and robust approaches, focusing on challenges such as generalization, scalability, offline learning, and learning from suboptimal or diverse experts. 
These directions bring imitation learning closer to real-world applicability. 
In this survey, we focus on this most recent era, highlighting and analyzing the latest contributions to the field.

\section{Explicit Imitation}
\label{section:explicit}

The explicit imitation setting assumes that expert demonstrations provide both the states visited and the actions taken. 
This setting is among the most straightforward forms of imitation learning, and serves as the foundation for many classic and modern methods. 
In this section, we organize the literature into two main strands: \textit{Behavioral Cloning} and \textit{Adversarial Imitation}. 
Behavioral Cloning has historically played a central role in this field, especially during the early adoption of deep learning. 
Recent works build upon Behavioral Cloning, enhancing its robustness or addressing its well-known weaknesses, as shown in Table~\ref{tab:bc} along with their objectives. 
In addition, a parallel line on explicit imitation research has evolved from adversarial learning frameworks, particularly following the introduction of {\em generative adversarial imitation learning (GAIL)}. This  line of work on {\em adversarial imitation}
has proposed various refinements or alternatives to improve GAIL stability and broaden its applicability
via addressing the challenges summarized in
Table~\ref{tab:adversarial}.

\subsection{Behavioral Cloning}

Behavioral Cloning (BC)~\cite{pomerleau1988alvinn,bain1995framework,ross2011reduction,daftry2016learning} is a relatively simple concept that applies supervised learning to imitation in one of its most straightforward forms. Like traditional supervised learning, it only requires a limited amount of offline data which, in this case, is a set of expert demonstrations. The agent learns by training on these demonstrations to directly map observations to actions, effectively mimicking the expert. Typically, these demonstrations consist of state-action pairs, which the agent uses to learn a policy. What sets BC apart from standard supervised techniques is that it applies supervised learning in a sequential decision-making context, where each action can influence future states. This makes BC a natural bridge between supervised learning and reinforcement learning (RL). One of its main challenges is covariate shift, the problem that arises when the agent encounters states during testing that were not present in the training data. Small errors can accumulate (compounding errors), leading the agent into unfamiliar territory where it performs poorly. Because of this, BC is often used as a pretraining method, providing a reasonable initial policy that can later be fine-tuned using other techniques~\cite{Masked_autoencoding,ramrakhya2023pirlnav}.

\begin{tabularx}{\textwidth} { 
   >{\centering\arraybackslash}X 
   >{\centering\arraybackslash}X }
   \caption{Works in Behavioral Cloning and Their Primary Objectives}\label{tab:bc}\\
    % \multicolumn{3}{|c|}{Country List} \\
    \hline
    Citation & Targeted Challenge \\
    \hline
    \hline
    ~\cite{hejna2023improving} & Covariate Shift\\
    ~\cite{ijcai2021p457} & Global Consistency\\
    ~\cite{pfss} & Generalization\\
    ~\cite{sasaki2021behavioral} & Expert Suboptimality\\
    ~\cite{10.1609/aaai.v37i9.26305} & Global Consistency\\
    ~\cite{yu2024usn} & Expert Suboptimality\\
    \hline
\end{tabularx}

The compounding errors issue, together with the lack of explicit planning or structure that is also not inherent in BC, makes it poorly suitable for long-horizon reasoning tasks, where small early mistakes can derail the rest of the trajectory.
To address this, Hejan {\em et al.} in~\cite{hejna2023improving} proposed a BC-based method that adds an instruction prediction objective to standard imitation learning. 
Expert demonstrations are still provided as state-action trajectories, but they are also accompanied with natural language instructions that describe the high-level intent behind the behavior (e.g., ``move to the door'').
During training, the model encodes the observations (or observation sequences) into latent states using a transformer encoder. It is then jointly trained to predict the next expert action and reconstruct the associated instruction using a sequence decoder.
Both predictions are supervised using ground truth of the expert: action prediction via cross-entropy, and instruction prediction via a language modeling loss. The total loss is a weighted combination of both.
At test time, the model does not predict instructions, it only uses the learned policy. 
The instruction prediction task is purely auxiliary, encouraging the encoder to learn goal-aware, structured representations, rather than just mimicking actions blindly.
This approach was shown in~\cite{hejna2023improving} to significantly improve generalization in long-horizon problems, and is also applicable to partially observed MDPs, where structure and memory are critical for success.

\begin{figure}[t]
  \centering
\includegraphics[width=0.9\textwidth]{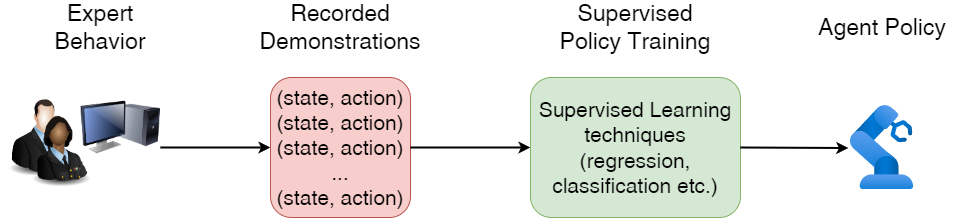}
\caption{Overview of BC~\cite{pomerleau1988alvinn,bain1995framework,ross2011reduction,daftry2016learning}. The expert performs the task, producing state-action data which is used to train a policy through supervised learning. The resulting policy allows the agent to imitate the expert's behavior.}
\Description{A flowchart showing the behavioral cloning process with four stages connected by arrows from left to right: expert behavior, recorded demonstrations (state-action pairs), supervised policy training (learning techniques like regression and classification), and learning agent policy.}
\label{fig:bc}
\end{figure}

Many behavior-cloning and sequential recommendation models optimize local correctness—accurate prediction of the next item—without explicitly enforcing global correctness, i.e., coherence of the entire interaction sequence. As a result, small one-step errors can accumulate into inconsistent sessions. 

It should be noted that BC
often 
optimize local correctness (accurate prediction of the next item), without explicitly enforcing global correctness, i.e., coherence of the entire interaction sequence.
This limitation becomes critical for sequential recommendation tasks, where predicting the next item a user will interact with requires capturing both short-term and long-term behaviors. 
In the work of~\cite{ijcai2021p457}, the authors propose a teacher-student design to tackle this problem by introducing three consistency-enhanced teacher models and one student agent. 
Each teacher is based on BERT4Rec~\cite{sun2019bert4rec} but is pretrained on a different self-supervised task: {\em temporal consistency} to ensure chronological ordering, {\em persona consistency} to preserve user-specific behavior patterns, and 
{\em global session consistency} to resist noise and maintain overall sequence coherence. 
These teachers are trained by manipulating raw sequential interaction data, using strategies such as shuffling, user-swapping, and mutual information maximization between local and global sequence parts. 
After pretraining, during student training, each teacher outputs both a prediction distribution and an internal representation of the sequence. 
The student model then learns by matching both the teachers' output predictions and their hidden representations through imitation losses based on KL divergence and MSE, respectively, allowing it to absorb consistency-aware knowledge beyond simple next-item prediction.
Although this method falls under the category of knowledge distillation, it is relevant to include in this survey as it directly addresses a limitation inherent to behavioral cloning.

Shifting from single-agent tasks to combinatorial optimization, Li {\em et al.}~\cite{pfss} apply imitation learning to the classic \textit{Permutation Flow Shop Scheduling} (PFSS) problem. 
PFSS involves determining the optimal permutation of jobs to be processed in the same machine order across multiple machines, with the goal of minimizing total completion time. 
The authors model the state as a combination of scheduled and unscheduled jobs, and define actions as selecting the next job to schedule. This removes the need for hand-crafted heuristics or expensive RL-based exploration, and instead proposes a lightweight, graph-based BC approach. 
The policy network is trained offline using a gated graph convolutional network~\cite{NIPS20172} to encode job features and an attention-based decoder~\cite{NIPS20171} to select the next job. 
The model is trained to imitate a heuristic scheduling algorithm, enabling it to learn job-ordering patterns that generalize to much larger problem instances. Their results demonstrate that imitation learning generalizes effectively to large-scale PFSS instances.

An important consideration in imitation learning, and in BC in particular, is the presence of suboptimality within expert demonstrations. 
Many existing approaches assume that demonstrations are fully optimal, although in practical settings, demonstrations often contain noisy or non-optimal behaviors. 
An extension to standard BC that aims to recover a policy capable of mimicking only the optimal behaviors, without requiring reward signals, environment interaction, or labels indicating demonstration quality is proposed in~\cite{sasaki2021behavioral}, extending the standard definition of IL (Definition~\ref{def}). 
The method begins by training an initial policy using standard BC, under the assumption that all demonstrated actions are equally valid. 
Afterwards, 
an
importance weighting mechanism is introduced into the training objective of the \textit{new} actual policy network. 
The
importance weights are not learned separately but are computed using the \textit{old} BC policy: for each state-action pair, if the old policy assigns a high probability to the action, it suggests high confidence and less noise contamination, leading to a higher weight. 
Lower confidence actions receive lower weights, reducing their influence during training. 
This weighted objective biases the new policy towards
imitating cleaner, more reliable behaviors. 
After each iteration, the newly trained policy is treated as the next ``old policy'', and the reweighting process is repeated multiple times to progressively refine the learned policy. 
During testing, the ensemble of final policies is averaged to further enhance robustness against residual noise. 

An extension of weighted BC, RelaxDICE, is proposed in~\cite{10.1609/aaai.v37i9.26305}. 
In this work, the authors aim to optimize over entire distributions induced by policies, learning a policy that behaves appropriately at a global level rather than focusing solely on local behavior. 
This is achieved by replacing the traditional importance weights with a density ratio that measures how frequently the new policy visits each state-action pair relative to the old behavior policy. 
By optimizing based on this density ratio, the method ensures that the learned policy remains within the support of the offline dataset while allowing for selective deviation toward expert-like behaviors, thus improving robustness against noisy demonstrations.

Finally, an approach that addresses suboptimal expert behavior, specifically targeting noisy action labels that can arise from low-quality annotators or faulty automated labeling systems, is presented in~\cite{yu2024usn}. 
The authors introduce \textit{Uncertainty-aware Sample-selection with Negative learning} (USN), which leverages the positive correlation between loss and uncertainty estimation to identify potentially noisy samples. 
By selecting large-loss samples that likely contain action noise and applying negative learning on complementary actions, USN corrects the bias introduced by noisy demonstrations. 
While initially developed for BC, the method demonstrates broad applicability, showing consistent improvements across adversarial approaches and offline methods as well, making it a versatile solution for robust imitation learning under diverse forms of action noise.

\subsection{Adversarial Approaches}

\label{GAIL}
\begin{figure}[t]
  \centering
\includegraphics[width=0.7\textwidth]{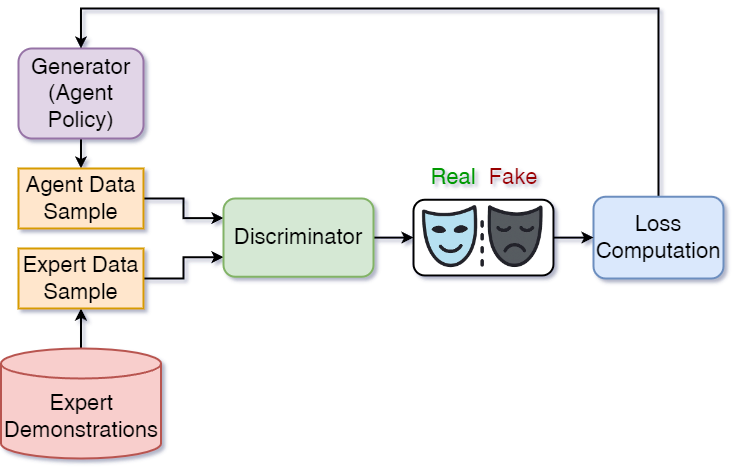}
\caption{Overview of GAIL~\cite{gail}. Expert and agent trajectories are fed to a discriminator trained to distinguish between the two. The agent updates its policy to fool the discriminator, leading to a policy that imitates the expert without access to reward or action labels.}
\Description{A flowchart diagram showing the GAIL architecture. At the top is a \textit{generator} (agent policy) box that feeds into an \textit{agent data sample} box. Below that is a \textit{expert demonstrations} database that feeds into an  \textit{expert data sample} box. Both sample boxes connect to a \textit{discriminator} in the center, which then connects to theater masks  representing the classification output (real vs fake), followed by a \textit{loss computation} box. There's also a feedback arrow from the generator back to the top, showing the adversarial training loop.}
\label{fig:gail}
\end{figure}

\textit{Generative Adversarial Imitation Learning} (GAIL)~\cite{gail} was introduced as a way to address the covariate shift problem inherent in BC. 
Drawing inspiration from \textit{Generative Adversarial Networks} (GANs)~\cite{goodfellow2014generative}, GAIL sets up a game between two components: a generator, which represents the agent’s policy attempting to mimic expert behavior, and a discriminator, which learns to distinguish between actions taken by the expert and those produced by the agent (Figure~\ref{fig:gail}). 
Throughout training, the generator is optimized to fool the discriminator into believing that its actions are indistinguishable from those of the expert, while the discriminator concurrently improves its ability to differentiate between the two. 
While training occurs online with the agent actively interacting with the environment, this process {\em does not} require access to the environment’s reward function. 
Instead, the generator $\pi(a|s)$ receives feedback in the form of: 
\begin{equation}
r(s,a)=-log\bigl(1-D\left(s,a\right)\bigr)   
\end{equation}
i.e., a reward derived from the discriminator's confidence $D(s,a)$, a measure of how convincingly %it 
the generator 
imitates the expert. This reward is used to improve the generator in a traditional RL way.
Meanwhile, the discriminator $D(s,a)$ is trained as a binary classifier, receiving feedback based on whether it correctly identifies the source of state-action pairs. 
Through this adversarial setup, the generator progressively approximates the behavior of the expert. 
By leveraging online interactions instead of relying solely on expert trajectories, GAIL allows the agent to generalize more effectively to novel states and recover from deviations, addressing one of the key limitations of BC.
Its success has since inspired a number of extensions and alternative adversarial imitation learning approaches, each refining different aspects of the framework.

\begin{tabularx}{\textwidth} { 
   >{\centering\arraybackslash}X 
   >{\centering\arraybackslash}X }
   \caption{Works on Adversarial Imitation Learning and their Primary Objectives}\label{tab:adversarial}\\
    % \multicolumn{3}{|c|}{Country List} \\
    \hline
    Citation & Targeted Challenge \\
    \hline
    \hline
    ~\cite{gail} & Covariate Shift\\
    ~\cite{wang2024exploring} & Stability\\
    ~\cite{li2017infogail} & Multi-Modality\\
    ~\cite{ijcai2020p405} & Multi-Modality\\
    ~\cite{Wang_Gao_Wu_Jin_Yao_Li_2023} & Data Privacy\\
    ~\cite{wang2021reinforced} & Data Privacy\\
    ~\cite{Ruan_Di_2022} & Stability\\
    ~\cite{sun2022deterministic} & Expert Suboptimality\\
    \hline
\end{tabularx}

One issue with the GAIL algorithm lies in how the reward is computed from the discriminator.
Its reward function $r(s,a)=-log\bigl(1-D\left(s,a\right)\bigr)$ behaves smoothly for low discriminator values, but as $D(s,a)\longrightarrow1$, the reward grows sharply. 
As a result, small changes in the discriminator’s output can lead to very large gradients, especially when the policy improves slightly.
This leads to unstable learning.
This problem is studied in detail in~\cite{wang2024exploring}, 
which provides a mathematical proof of the fact that gradient explosion is a provable risk in GAIL when using deterministic policies.
Since these policies lack exploration, their updates are especially vulnerable to large, destabilizing gradients. If the discriminator becomes overconfident early in training, a common scenario, it can assign near-one scores to poor transitions, causing massive reward spikes and unstable updates.
To fix this, the authors propose a simple yet effective solution called \textit{Clipping REward of Discriminator Outliers} (CREDO). The idea is to clip the reward at a maximum threshold $\delta$, modifying the reward function to:
\begin{equation}
    r(s,a)=min\left(-log\bigl(1-D\left(s,a\right)\bigr),\delta\right)
\end{equation}
This prevents the reward from becoming unbounded, stabilizing the policy updates while still giving higher scores to expert-like behavior.

A key limitation of GAIL is its inability to handle multi-modal expert demonstrations. 
In many real-world settings, there are multiple valid, distinct strategies for acting in the same state (e.g., a car approaching a slower vehicle may choose to pass on the left or the right, or a robot may grasp an object from above or from
the side). 
GAIL assumes all demonstrations come from a single behavior mode, leading to {\em mode collapse}: the learned policy averages over diverse strategies, producing blurry, unrealistic behaviors that 
cannot possibly 
correspond to 
any expert ones.
\textit{InfoGAIL}~\cite{li2017infogail} addresses this by extending GAIL with a latent variable $c$, which acts like a behavior selector or mode switch. 
For example, $c$ might distinguish between aggressive vs. cautious driving, or between left vs. right lane changes. 
These codes are unlabeled and InfoGAIL only requires specifying the number of behavior modes (i.e., the dimensionality of $c$). 
The policy $\pi$ is then conditioned on this
variable---i.e., we have $\pi(a|s,c)$---enabling controllable and interpretable behavior based on the behavior $c$ we input.
To ensure the policy actually uses $c$, InfoGAIL introduces a mutual information objective. 
It trains a network $Q(c|s,a)$ (not to be confused with the $Q$-learning function), that in its simplest form is a classifier, to recover the latent code used to generate a given trajectory. 
During training, the agent samples a random code $c$, generates a trajectory using $\pi(a|s,c)$, and uses that same $c$ as the target for training $Q$. 
This setup, after a while encourages the policy to produce distinct and consistent behaviors for each code.
Meanwhile, the discriminator still learns to distinguish expert from agent-generated behavior, 
and is
agnostic to $c$. 
The result is a policy that not only mimics expert demonstrations, but also discovers and separates different strategies within them, without requiring any labeled expert data.

Another method that extends GAIL to address the challenge of multi-modal expert behavior is \textit{Triple-GAIL}~\cite{ijcai2020p405}. 
Unlike InfoGAIL, which learns behavior modes in an unsupervised manner but still requires manual specification of a latent strategy $c$ at test time, Triple-GAIL introduces a mechanism for automatically selecting the appropriate behavior strategy during execution. 
This is made possible with the assumption that the expert demonstrations are labeled with their corresponding strategies.
Triple-GAIL builds upon the GAIL framework by introducing a third component. 
In its architecture, both the generator $\pi(a|s,c)$ and the discriminator $D(s,a,c)$ are conditioned on the {\em skill label} $c$, which represents the expert's strategy. 
The key addition is the selector $C(c|s,a)$, a model trained to predict the strategy $c$ based on the observed state-action pair.
This allows the agent to adaptively switch strategies depending on the current context. 
For example, if a vehicle is overtaking another from the left but an obstacle suddenly appears ahead, it may be safer to pass on the right instead. 
Triple-GAIL enables this kind of adaptive behavior without manual intervention.
The selector is trained using both expert demonstrations (to match ground-truth labels) and agent-generated experiences (to improve generalization and robustness). 
During training, the generator and selector jointly compete against the discriminator in a three-player adversarial setup, learning to produce realistic and context-appropriate behavior that aligns with expert strategies.

One 
capability of GAIL
that was only recently explored, 
is its potential to generate data without breaching the privacy of its sources, particularly because it relies on indirect training signals rather than direct access to supervision as in BC. 
In~\cite{Wang_Gao_Wu_Jin_Yao_Li_2023}, the authors address the task of generating synthetic human mobility trajectories while preserving the privacy of the users whose data is utilized during training with the introduction of PateGAIL. 
Since mobility trajectories can reveal sensitive personal information, collecting large-scale real-world datasets creates serious privacy concerns. 
To address this, the authors employ GAIL in a modified setting. 
While the generator is trained on a central server that communicates with users, each user's discriminator is trained locally, 
determining
the realism of transitions based only on private user-specific data. 
This ensures that no raw trajectories are uploaded to the server. 
Instead, the generator receives only reward signals from the discriminators, allowing it to be guided toward realistic behavior without accessing any private data. 
Further, to 
guarantee differential privacy, Laplace noise is added to the rewards, mitigating the possibility that the transmitted rewards could still leak private information. 
To complete its guarantees for the creation of privacy-preserving synthetic human mobility data, the paper proposes that the generator is trained with federated learning techniques.

Another approach to synthetic data generation is the construction of user profiles. 
User profiles have numerous applications including predicting future behaviors, supporting recommendation systems, and enabling urban planning and targeted services.
For instance,~\cite{wang2021reinforced} 
puts forward a system to learn
mobile user profiles 
based on mobility behavior without addressing privacy concerns. 
Their objective is not to predict 
a
next ``point of interest'' (POI) 
directly, but to build a profile that can effectively summarize user behavior. 
To achieve this they use an adversarially-inspired framework, though not in the traditional GAIL setting. 
Two modules are involved in the algorithm, a representation module and a imitation module.
The representation module acts as a generator and outputs a vector representation of the user profile. 
To evaluate the quality of this profile during training the authors introduce an imitation task where a DQN-based imitation module predicts the user's next point of interest given the generated profile. 
The prediction is evaluated through a joint reward that combines a distance reward measuring physical POI distance, a category reward measuring similarity of POI types, and a match reward checking for exact location match. 
The imitation module is trained using standard DQN techniques via the Bellman equation while the representation module is updated adversarially to maximize the expected reward. 
The imitation module acts solely as a training signal to refine the profiles and only the representation module is used at test time to provide the final user profile.

 In many real life domains, expert demonstrations are influenced by hidden variables that cannot be logged or are subject to random variation. These hidden variables, known as \textit{unobserved confounders}, affect the observed inputs (e.g., states and actions) and the outcome (rewards). 
~\cite{Ruan_Di_2022} take a causal inference route that explicitly models and addresses unobserved confounders 
when 
imitating human driving behavior.
Specifically,
the driver's mood or the weather conditions are unobserved confounders and may impact both the policy of the expert driver and the returned reward of the environment.
These factors
create misleading links between what the agent observes and what it tries to replicate. In the context of causal reasoning, these links are referred to as $\pi$-backdoors and are typically undesirable. Previous work in IL either ignored confounders entirely or relied on fully observable variables. 
By contrast, to
 address this
 issue, the authors propose \textit{Sequential Causal Imitation Learning} (SeqCIL), a framework that extends the GAIL model by incorporating causal reasoning into the discriminator's objective. While the generator policy network remains structurally unchanged, the discriminator is trained to block $\pi$-backdoor paths by learning from trajectory segments that reflect causal dependencies. This discriminator provides a reward signal to the generator that reflects causal correctness.

While adversarial imitation approaches like GAIL are powerful, they can suffer from sample inefficiency, instability due to min-max optimization, and non-stationary reward functions. 
To address these issues, ~\cite{sun2022deterministic} proposes \textit{D2-Imitation}, a non-adversarial alternative that is better suited for deterministic policies and is significantly easier to train.
Rather than copying actions via supervised learning as in BC, D2-Imitation still trains a policy by optimizing rewards in a similar way to GAIL; however, the reward is {\em not} learned adversarially. 
Instead, D2-Imitation introduces two replay buffers: $B^+$ for storing expert-like transitions (initialized with expert transitions) and $B^0$ for storing non-expert transitions (initialized with random generated transitions).
The discriminator is pretrained using those two initialized buffers and is not being further updated during training. This avoids the instability that arises from adversarial reward shaping and ensures a more stable training process.
Once the policy begins generating new transitions during training, the discriminator evaluates each and routes it into either buffer based on whether its confidence exceeds a threshold.
Transitions in $B^+$ are assigned a reward of $1$, and those in $B^0$ receive a reward of $0$. The policy is then trained via TD-learning to maximize the likelihood of producing high-reward transitions.
While D2-Imitation lacks the continuous reward shaping of GAIL and may be less expressive in some settings, it is more robust in low-data or high-variance scenarios where GAIL can collapse.

\section{Implicit Imitation Learning}
\label{section:implicit}
In implicit imitation, the learning agent has access only to sequences of expert states or state transitions, with no information about the expert's actions. 
This less constrained but more challenging setting has led to a broad range of approaches. 
In this section, we group existing approaches into two general categories: model-based and model-free methods. 
Earlier research in this area often relied on building inverse dynamics models to infer missing action information, forming the core of model-based strategies. 
Although this line of work is limited in terms of attention lately, new works still emerge (Table~\ref{tab:modelbased}). 
On the other hand, model-free techniques have gained substantial attention in recent years (Table~\ref{tab:modelfree}). 
These approaches typically involve learning policies or objectives directly from observations, often integrating ideas from reinforcement learning or adversarial training to support learning in action-free settings.

\subsection{Model-Based}

In the model-based context, a direct successor to behavioral cloning for state-only expert datasets is \textit{Behavioral Cloning from Observation} (BCO)~\cite{ijcai2018p687}. 
Since the expert's actions are unavailable in this setting, a direct supervised learning approach is not applicable. 
To address this, BCO begins with a pre-demonstration exploration phase in which the agent interacts with the environment using a random policy and collects transitions that map actions to resulting state changes. 
This data is then used to train an inverse dynamics model that predicts the most likely action given a state transition. 
Once this model is learned, the expert’s state-only trajectories are parsed, and the corresponding missing actions are inferred. 
Standard behavioral cloning is then applied using the inferred state-action pairs to train a policy that approximates the expert’s behavior. 
An extended variant of this method proposed in the same work, BCO($\alpha$), introduces post-demonstration interaction, enabling the agent to further refine its inverse model and policy. 
In this iterative setting, the current imitation policy is executed to collect new transitions, which are then used to update the inverse model, re-infer the expert’s actions, and retrain the policy accordingly. 
By repeating this cycle over multiple iterations controlled by the parameter $\alpha$, the model gradually improves its accuracy, resulting in a more effective imitation policy.

\begin{tabularx}{\textwidth} { 
   >{\centering\arraybackslash}X 
   >{\centering\arraybackslash}X }
   \caption{Works in Model-Based Implicit Imitation Learning and Their Primary Objectives}\label{tab:modelbased}\\
    % \multicolumn{3}{|c|}{Country List} \\
    \hline
    Citation & Targeted Challenge \\
    \hline
    \hline
    ~\cite{GONZALEZ2022117167} & Application\\
    ~\cite{household} & Generalization\\
    \hline
\end{tabularx}

An approach that focuses on learning behavioral models through implicit imitation instead of creating new policy behaviors is presented in~\cite{GONZALEZ2022117167} and focuses on human vehicle driving.
The goal is to identify whether or not an \textit{Attention Deficit/Hyperactivity Disorder} (ADHD) afflicted driver is showing attention deficit related risky behaviors. 
The authors investigate real driving data, observations only, from two human groups consisting of medicated ADHD drivers and unmedicated drivers. 
For each group they train a separate network using a standard BC regression method, tasked with predicting the next state based on the current input state. 
After training, both models are capable of predicting future states given scenarios representative of their respective behaviors. 
During testing the system inputs observed states into both models and collects the corresponding predictions. 
The predicted next states are then compared against the actual observed next states and error metrics are calculated. 
A lower prediction error indicates that the driver's current behavior is closer to the behavior represented by the corresponding model, allowing the system to infer whether the driver is acting under medicated or unmedicated conditions.

While most implicit imitation works focus primarily on low-level behavior, moving away from action-level learning towards semantic-level modeling is sometimes necessary to better guide a robot. 
In~\cite{household}, the authors focus on semantic task model extraction, integrating common sense into a household service robot designed to perform human tasks such as setting a table, cleaning up, and related activities. 
Their goal is to design a plan sequence that can adaptively complete a given task, using only observations. 
Training begins by detecting key changes in object states and identifying predefined semantic actions such as ``move'', ``open'', and ``grasp''. 
For each action segment observed, the system learns its preconditions by capturing relevant environment information and object states before the action, and its post-conditions based on the resulting state after the action. 
These relationships are mapped into a task graph, with action segments as nodes and edges connecting actions whose preconditions are satisfied by the postconditions of preceding actions, forming a causal action dependency graph. 
Afterwards, the system applies common-sense constraints to filter physically implausible or illogical action sequences. 
These include posture constraints which ensure ergonomic and physically feasible human-like movements, grasp type constraints which match appropriate grasp styles to pick an object, and tool-environment interaction constraints which ensure correct object placement such as placing items on stable surfaces rather than floating in midair. 
At test time, given a current world state, the agent identifies which action preconditions are satisfied and moves forward by suggesting a sequence of predefined commands for the robot to execute.

\subsection{Model-Free}

The first model-free implicit imitation learning approach was in fact an adversarial one, directly inspired by the original GAIL algorithm (see Section~\ref{GAIL})~\cite{torabi2018generative}.
The main premise and concept of the approach remain the same as GAIL's---meaning that the goal is to imitate an expert while interacting with the environment to avoid compounding errors, and this is achieved in an adversarial manner by using a discriminator to identify whether a transition is likely to come from the expert. 
The key difference is that the discriminator no longer considers state-action pairs, but instead focuses on state and next-state pairs. 
This modification breaks the traditional dependence on inverse dynamics models that was previously necessary to address learning from observation problems and allows for a direct, model-free imitation method that scales effectively to both low-dimensional features and high-dimensional raw visual inputs.

\begin{tabularx}{\textwidth} { 
   >{\centering\arraybackslash}X 
   >{\centering\arraybackslash}X }
   \caption{Works in Model-Free Implicit Imitation Learning and Their Primary Objectives}\label{tab:modelfree}\\
    % \multicolumn{3}{|c|}{Country List} \\
    \hline
    Citation & Targeted Challenge \\
    \hline
    \hline
    ~\cite{torabi2018generative} & Covariate Shift\\
    ~\cite{wang2024diffail} & Stability\\
    ~\cite{wu2021textgail} & Text Generation\\
    ~\cite{pmlr-v229-wang23a} & Global Consistency\\
    ~\cite{NEURIPS2021_868b7df9} & Generalization\\
    ~\cite{liu2024imitation} & Generalization\\
    ~\cite{chrysomallis2023deep} & Expert Suboptimality\\
    ~\cite{Chrysomallis_Chalkiadakis_Papamichail_Papageorgiou_2025} & Expert Suboptimality\\
    \hline
\end{tabularx}

A key limitation in GAIL is that its discriminator is a simple binary classifier, meaning it only learns to say whether a transition is from the expert or not. 
This binary setup can provide sparse and unstable reward signals, and fails to capture the full expert distribution. 
To address this,~\cite{wang2024diffail} propose DiffAIL, a novel extension that replaces the discriminator with a diffusion model.
Diffusion models work by learning to reconstruct samples from noise through a gradual denoising process (e.g., turn a noisy image back into a high-quality photograph).
In the case of DiffAIL, the diffusion model is trained to reconstruct expert demonstrations that were corrupted with Gaussian noise.
During training, the diffusion model is first trained offline on expert data only, using a denoising score-matching objective. 
Once trained, it serves as a fixed discriminator. When the agent generates a new transition, noise is added to that transition and the diffusion model tries to reconstruct it. 
The error in reconstruction is interpreted as how ``non-expert-like'' the transition is, with a smaller error yielding a higher reward, and a larger error giving a lower reward. 
This leads to a reward function that is continuous, dense, and well-aligned with the expert distribution.
DiffAIL demonstrates strong performance across multiple benchmarks, outperforming GAIL and other recent methods in both the state-action setting (explicit imitation) and the state-only setting (implicit imitation). 

TextGAIL, a framework that applies GAIL to text generation tasks is intoduced in~\cite{wu2021textgail}. In this context, the policy network becomes a language model, which treats tokens as discrete actions. At each time step, the model takes the current sentence prefix as input (the “state”) and outputs the next token (the “action”), thereby forming a full sequence one step at a time.
By introducing GAIL to this framework, they reduce variance and improve performance of the model, giving rewards without hand-engineering and focusing on whole-text quality.
To enable GAIL in the language domain, several adaptations are introduced. 
First, instead of a standard GAIL policy trained on continuous state-action pairs, the generator is a pretrained language model (like GPT-2) trained with proximal policy optimization (PPO)~\cite{schulman2017proximal} to stabilize the high-variance gradient updates typical in text generation. 
Second, the discriminator is replaced by a contrastive discriminator. Rather than evaluating one sentence at a time, it compares an expert and a generated sentence jointly and learns to score them in relative terms.
The generator and discriminator are trained in alternating steps using a replay buffer that mixes human-written sequences with generated ones.
While the generator produces discrete actions (tokens), there are no explicit expert action labels per time step, and the reward is inferred from full trajectories via adversarial training. 
Thus, although the training mechanism resembles explicit imitation learning, the data access and structure place TextGAIL in the category of implicit imitation learning.

An implicit imitation approach for long-horizon imitation learning is introduced by~\cite{pmlr-v229-wang23a}. 
The authors propose a hierarchical framework that splits the problem into planning and control. 
The planning component is trained using human play videos, which capture the sequence of subgoals needed to accomplish a task, while the control component is learned from a small number of robot teleoperation demonstrations to achieve precise low-level manipulation. 
Unlike prior works that rely heavily on dense teleoperated demonstrations~\cite{mandlekar2020iris,Mandlekar-RSS-20,shiarlis2018taco}, this approach leverages cheap, unlabeled human play data and a small set of robot demos. 
For the human-driven component, around 10 minutes of hand interaction per environment are recorded using two cameras to reconstruct 3D hand trajectories, without requiring action labels or rewards. 
This contrasts with traditional teleoperation-heavy pipelines. 
These visual sequences are encoded into embeddings, and a Gaussian Mixture Model is trained to represent diverse latent plans that capture multimodal strategies toward reaching goals. 
On the robot side, a transformer-based policy network is trained via behavior cloning, conditioned on the latent plans. 
To align the human and robot visual domains, a KL divergence regularization is applied between their embedding distributions. 
At test time, given a current and goal observation, the planner outputs a latent plan that guides the robot controller to produce executable actions.

Moving away from adversarial approaches, an alternative method that trains policies through reinforcement learning while defining its own reward function purely from observations is proposed in~\cite{NEURIPS2021_868b7df9}. 
The main idea involves learning a policy based on how close a given state is to the task goal. 
A proximity function $f(s)$ is defined to measure how close the state $s$ is to the goal, where higher values indicate greater proximity, and this function is pretrained using only the observation data provided by the expert. 
During training, after the agent takes an action and reaches a new state $s'$, the reward is 
computed as $r(s,s')=f(s')-f(s)$. 
Positive rewards indicate progress toward the goal, while negative values indicate divergence. 
The policy is then trained to maximize cumulative rewards using a PPO framework~\cite{schulman2017proximal}. 
This formulation allows the policy to generalize effectively to unseen states, as it learns to minimize the distance to the goal rather than simply imitating expert actions.

A similar approach is presented in~\cite{liu2024imitation}, which also defines a reward function based solely on observations and trains a policy in a deep reinforcement learning fashion. 
Instead of designing the reward based on the distance to a final goal, they directly compare entire state trajectories to derive rewards. 
During training, rather than relying on individual transitions, the agent acts through a full episode. 
Afterwards, the reward for the trajectory is computed by comparing the distribution of the agent's states to that of the expert using \textit{Optimal Transport} (OT)~\cite{peyre2019computational}. 
This matching allows the agent to obtain a clear signal of how far its trajectory deviated from expert behavior without being constrained by strict temporal alignment, and to update its policy accordingly. 
The main focus of this work, however, is the introduction of discount $\gamma$ scheduling. 
Rather than keeping the discount factor fixed throughout training, they propose a scheduler that gradually increases its value. 
This strategy enables stable early-stage learning focused on short-term behavior and progressively incorporates long-term planning as the policy improves. 
If the discount factor is set high from the beginning, training becomes unstable and noisy. 
To automate the scheduling, they monitor the agent's value estimates during training and adjust the discount factor upward based on observed improvements, enabling automatic progression toward long-horizon behavior learning without manual tuning.

\begin{figure}[t]
  \centering
\includegraphics[width=0.7\textwidth]{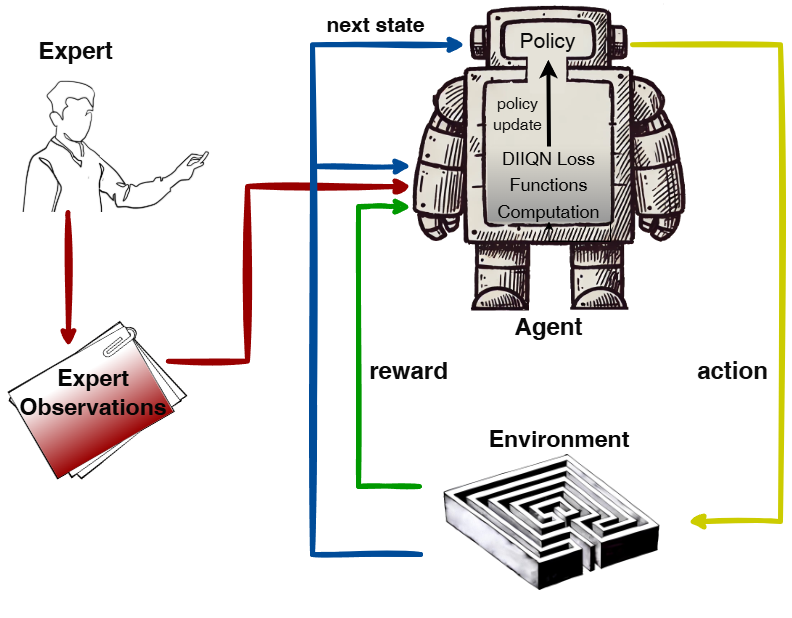}
\caption{High-level visualization of DIIQN~\cite{chrysomallis2023deep}, which combines standard deep reinforcement learning with imitation learning by optimizing two loss functions, one reflecting expert behavior and one guiding the agent’s own policy.}
\Description{A system diagram showing the DIIQN framework. The agent in the center interacts with an environment (shown as a maze-like structure at the bottom) through arrows: \textit{action} output, \textit{next state} input, and \textit{reward} feedback, forming a complete reinforcement learning loop. Additionally, an expert provides observations that feeds into the robot agent. The agent performs policy updates through the DIIQN loss functions computation components. }
\label{fig:diiqn}
\end{figure}

A distinct approach that leverages imitation learning in combination with DRL is the one introduced in~\cite{chrysomallis2023deep}, focusing on observations only, through the \textit{Deep Implicit Imitation Q-Network} (DIIQN) framework. 
Imitation learning is a relatively cheap and fast way to acquire a behavioral model or policy compared to DRL, but its performance is often capped by the expert’s skill level. 
By contrast, DRL enables convergence to optimality, but typically requires costly and computationally intensive exploration. 
The DIIQN method combines these two approaches by integrating implicit imitation into an online, model-free DQN-based pipeline without relying on expert action labels  (Figure~\ref{fig:diiqn}). 
This goes beyond the standard application of IL of basic mimicry (Definition~\ref{def}) and aims to adapt in situations where expert optimality is not guaranteed.
Training starts with a warmup phase, during which the agent explores randomly and builds an initial mapping between its actions and expert state transitions. 
During learning, the agent interacts with the environment and receives rewards at each step. 
However, for every transition, it also stores the closest expert transition along with an estimated action, which is continually refined through experience. 
When training the model, two loss functions are computed: one based on the agent’s own experience and one based on the expert-guided transition reconstructed from the agent’s perspective. 
The loss resulting in smaller policy value divergence is selected to guide the policy update, promoting stability during training. 
Unlike most imitation approaches, DIIQN does not replace the environment reward with an imitation-derived reward signal but continues to train solely based on the native environment rewards while using expert observations to guide and accelerate the learning process. 
This allows the agent to benefit from the expert’s guidance in the early stages to accelerate progress, while gradually refining its policy independently and potentially surpassing non-optimal experts. 
As a result, DIIQN significantly speeds up training compared to standard DRL techniques and leads to performance competitive with or exceeding that of the expert.

An immediate extension of DIIQN tailored to heterogeneous scenarios is Heterogeneous Actions DIIQN (HA-DIIQN)~\cite{Chrysomallis_Chalkiadakis_Papamichail_Papageorgiou_2025}.
This method addresses the case where the expert and agent operate under different action spaces, removing the restrictive assumption that the agent can replicate the expert’s actions. 
Such heterogeneity arises frequently in practical applications, such as robotics, where agents may differ in degrees of freedom or actuator capabilities. 
Notice that this work deviates from the standard IL assumption (cf. Definition~\ref{def}) where the agent adopts the expert's action space. 
Instead of pursuing the expert policy $\pi_e: S \rightarrow \Delta(A)$, they seek to learn a policy $\pi: S \rightarrow \Delta(A_a)$ that operates over the agent's distinct action space $A_a$.
The proposed approach begins by identifying expert transitions that are infeasible for the agent, using a KL-divergence-based criterion to check whether a similar transition has been encountered during the agent’s own experience. 
If no such match exists, the transition is considered infeasible. 
In those cases, the method attempts to discover an alternative sequence of agent experiences that intersects with the expert’s future trajectory. 
This process, termed \textit{k-n step repair}, searches up to $k$ steps within the agent’s experience replay and up to $n$ steps along the expert’s demonstration, with these parameters chosen based on the complexity of the task. 
When a valid bridging path is found, the expert-guided loss function is constructed using the first transition of the bridge in place of the infeasible expert transition. 
If no bridge is available, no expert information is used for that particular update, and only the agent's original loss is applied. 
The selection of the final loss function for training follows the same divergence-based mechanism as in the original DIIQN framework, thereby preserving its advantages while extending its applicability to heterogeneous action settings.

\section{Inverse Reinforcement Learning}
\label{section:irl}

\textit{Inverse Reinforcement Learning} (IRL) is a subcategory of imitation learning that shifts the focus to a slightly different question. 
Rather than asking how to replicate the expert's behavior, it attempts to uncover the reason behind that behavior, effectively answering the ``why'' question. 
In practice, this means that an IRL algorithm aims to infer the {\em reward function} that
guides expert behavior, using %only 
observed expert demonstrations~\cite{ng2000algorithms}. 
This is especially useful in scenarios such as autonomous driving, where manually designing a reward function is difficult, but examples of expert behavior are readily available and already reflect the rules and preferences we wish to encode. 
Once the reward function is derived, it can typically be used to train a policy using any DRL method of one's choosing. Beyond that, it can help align agents with human preferences~\cite{deshpande2025advances}, enhance behavior explainability~\cite{xie2022towards}, and infer individual goals in multi-agent scenarios to support better planning~\cite{deshpande2025advances}.
Moreover, as cited in~\cite{ghavamzadeh2015bayesian}, IRL can also be valuable for constructing computational models of animal and human learning processes, as well as for modeling opponent behavior in game-theoretic settings.

Although methods like GAIL and other GAIL-inspired approaches discussed in Section~\ref{GAIL} are often grouped under the umbrella of IRL, we choose to distinguish them from the works presented in this section. 
GAIL trains a discriminator to distinguish between expert and agent behavior, and uses this signal to improve the agent's policy. 
While this process serves a similar role to inferring a reward function, it does so implicitly. 
In the literature, this resemblance is often enough to categorize such methods as IRL. 
However,
IRL traditionally aims to
explicitly recover a reward or cost function; 
as such, adversarial IL and IRL approaches occupy distinct spots in our taxonomy. 

\begin{tabularx}{\textwidth} { 
   >{\centering\arraybackslash}X 
   >{\centering\arraybackslash}X }
   \caption{Works in Inverse Reinforcement Learning and Their Primary Objectives}\label{tab:irl}\\
    % \multicolumn{3}{|c|}{Country List} \\
    \hline
    Citation & Targeted Challenge \\
    \hline
    \hline
    ~\cite{kumar2023graph} & Third-Person Expert\\
    ~\cite{zhou2024rethinking} & Expert Suboptimality\\
    ~\cite{NEURIPS2022_d842425e} & Data Privacy\\
    ~\cite{swamy2023inverse} & Efficiency\\
    \hline
\end{tabularx}

The first approach to inverse reinforcement learning in the deep learning era was introduced by~\cite{pmlr-v48-finn16}, which focused on learning a cost function instead of a reward function. 
While a reward function assigns higher values to desirable behavior, a cost function assigns lower values to the same, making it a mirror formulation where minimizing cost is equivalent to maximizing reward. 
In this method, a neural network is trained to represent the cost function by producing low values for expert demonstrations and higher values for other, non-expert trajectories. 
Since the expert is assumed to act in ways that minimize some internal notion of cost, their behavior is treated as optimal under this cost function, and deviations from it are interpreted as less optimal or more expensive. 
The policy is trained to minimize the learned cost, and new data is collected from the updated policy. 
The cost function is then refined using this data alongside the original expert demonstrations. 
This loop of alternating between cost learning and policy optimization continues until the cost function converges and the policy reliably imitates expert behavior.

A new way of learning from video data that avoids operating directly on raw pixel features is presented in~\cite{kumar2023graph}. 
Instead of relying on noisy visual input that can create a complex and distracting learning problem for the agent, a graph-based approach is introduced to focus on the underlying object interactions and use them to construct a reward function. 
The method analyzes third-person expert demonstrations by first applying a pretrained object detector~\cite{russakovsky2015imagenet}, treating each detected object as a node in a graph, with edges representing potential interactions between object pairs. 
For every frame, a fully connected graph is constructed, and an \textit{Interaction Network}~\cite{battaglia2016interaction} is used to encode the relationships into a single graph embedding that summarizes the state of the scene. 
This embedding captures the structural configuration of objects in the task, regardless of their visual appearance in the original video. 
The goal is to create a reward function that reflects how far an agent has progressed through a task based on the current scene configuration. 
By processing multiple expert videos and comparing their embeddings, the method aligns frames based on shared task structure, under the assumption that all experts are solving the same task. 
This temporal alignment is learned using \textit{Temporal Cycle Consistency}~\cite{dwibedi2019temporal}, which encourages frames at similar stages of progress across different videos to be mapped close together in embedding space, even when the videos differ in appearance or timing. 
As a result, the method builds a representation of task progression that acts like a progress bar, where each frame's position in the embedding space indicates how close it is to completing the task. 
The final reward function is defined as the distance between the current frame's embedding and that of the goal frame, allowing reinforcement learning to proceed based on how much progress has been made toward task completion.

Moving to the problem of expert suboptimality, it also remains a central concern in inverse reinforcement learning as well. In~\cite{zhou2024rethinking}, the authors address the limitations of assuming globally optimal experts by proposing an approach that learns reward functions from demonstrations of varying quality. 
Rather than matching the expert’s behavior directly, their method uses a ranked set of trajectories $\tau$, such that $\tau_i\prec\tau_j\prec...$, meaning $\tau_j$ is preferred over$\tau_i$. 
This ranking serves as the key supervisory signal for reward learning, allowing the model to move away from strict data alignment toward task alignment. Given these ranked preferences, they train a neural network reward function $r$ with parameters $\theta$ such that
\begin{equation}
    \sum_{s \in \tau_i} r(s;\theta) < \sum_{s \in \tau_j} r(s;\theta), \text{for $\tau_i\prec\tau_j$}
\end{equation}
The reward function is trained using a pairwise ranking loss~\cite{brown2019extrapolating} that encourages trajectories higher in the ranking to accumulate more reward than those lower in the ranking. 
This does not imply that the reward function assigns maximal reward to the top-ranked trajectory, but rather that it captures the underlying structure of task success implied by the ordering. 
In doing so, the method avoids the need for globally optimal expert behavior and instead leverages relative performance to guide learning.

An interesting approach that handles multiple experts in a distributed and privacy-preserving manner is presented in~\cite{NEURIPS2022_d842425e}. 
Instead of combining all expert demonstrations into a single dataset for centralized training, the authors propose splitting the data across multiple learners, with each learner receiving a different subset. 
Each learner has access only to its own data, creating a setting where multiple agents aim to infer the same reward function despite observing different demonstrations. 
At certain steps, learners communicate with one another to share parameter estimates and collaboratively refine their understanding of the underlying reward function. 
In addition to reward learning, the method also infers constraint cost functions, capturing state-action pairs that the experts try to avoid. 
Through an alternating optimization process, the agents iteratively update both reward and constraint parameters, eventually agreeing upon a shared policy that generalizes across the distributed data.

An important aspect of IRL is what follows after a reward or cost function has been inferred.
In most IRL works, it is either stated explicitly or implied that once the reward function is learned, it can be used to train a policy using conventional reinforcement learning. 
However, 
reinforcement learning can be computationally expensive and sample inefficient~\cite{kostrikov2021improving,yin2022planning,hu2023imitation}. 
In~\cite{swamy2023inverse}, the authors propose an alternative that avoids costly RL training by leveraging the available expert demonstrations more directly. 
While their objective differs, their training mechanism shares a conceptual similarity with Go-Explore~\cite{ecoffet2021first} in the way the agent revisits meaningful states. 
During training, the agent is repeatedly reset to states visited by the expert, then allowed to act under its current policy. 
The resulting rollouts are compared against the expert’s future behavior from the same states, and the differences are used to update both the policy and the reward function. 
However, as with Go-Explore, this approach suffers from the assumption that the environment is simulated or can be reset to arbitrary expert states, which may limit its applicability in real-world settings.

\section{Challenges and Future Endeavors}
\label{section:challenges}

As highlighted throughout this survey, imitation learning is a well-explored and mature field, yet ongoing research continues to address its limitations and unresolved challenges. Below, we summarize some of the most pressing open problems.\\

\textbf{Covariate shift} is one of the most prominent issues, which, despite significant improvements, still constrains performance in many settings. Recent adversarial approaches have made progress in mitigating this problem, but they often introduce new challenges such as training instability and sensitivity to hyperparameters.

\textbf{Access to optimal expert data} is another challenge, as high-quality demonstrations are rarely guaranteed in complex, real-world scenarios. Although recent work has begun to address learning from suboptimal data, many existing methods still rely heavily on the assumption of expert optimality, which limits their robustness and applicability.

\textbf{Global consistency} refers to the difficulty of maintaining coherent, goal-directed behavior across long horizons or diverse tasks. This remains an open challenge, particularly in settings where agents must plan over extended sequences or adapt to varied situations.

\textbf{Multi-modality in expert data} presents further difficulties. When demonstrations reflect multiple valid strategies, agents must learn to differentiate between them rather than averaging behaviors into a single, incoherent policy.

\textbf{Data privacy and ethical concerns} are beginning to emerge as important considerations, especially in applications involving sensitive or proprietary demonstrations. However, these issues are only marginally addressed in the current literature.\\

In addition to these, several other key challenges remain underexplored.\\

\textbf{Safety} is a critical but often overlooked aspect, particularly in high-stakes domains such as autonomous driving and healthcare. While certain approaches, such as ``guardian'' models~\cite{peng2022safe}, have attempted to incorporate safety constraints, the field lacks comprehensive mechanisms to ensure safe behavior during training and deployment.

\textbf{Demonstration cost and data efficiency} are also pressing concerns. Collecting high-quality demonstrations can be expensive or infeasible, and many imitation learning algorithms require large volumes of data to perform effectively, especially in high-dimensional or long-horizon environments.

\textbf{Multi-agent imitation learning} remains largely unexplored, despite extensive theoretical foundations in multi-agent systems. Only a few recent works have begun to address the unique challenges posed by multi-agent settings.

Finally, \textbf{the lack of standardized evaluation practices} makes it difficult to compare methods objectively. Most approaches are tested in bespoke environments designed to showcase specific strengths. The field would benefit greatly from common benchmarks and evaluation protocols to enable fair and reproducible assessments.

\section{Conclusions}
\label{section:conclusion}

In this paper, we provided a comprehensive overview of recent advancements in the field of imitation learning, showcasing and explaining in detail a wide range of modern algorithms with a particular focus on the deep learning era.
We began by discussing core concepts and questions relevant to the domain, introducing important terminology and highlighting notable applications of imitation learning research.
We then presented the historical evolution of the field, tracing its development from the $1980$s to the present, and explaining why a new review is timely in light of recent progress.
To support this goal, we proposed a new taxonomy inspired by previous structures and aligned with current trends in imitation learning. 
Based on the established frameworks and advancing research directions, our taxonomy distinguishes between explicit, implicit, and inverse reinforcement learning approaches, and further categorizes each into subdomains. 
We then conducted an in-depth review of representative works in each category, analyzing their methodologies, assumptions, and contributions.
While substantial progress has been made in recent years, we identified several core challenges and limitations in existing approaches. 
We concluded by highlighting these issues and outlining what we believe are promising directions for future research.
We hope this survey will serve as a foundation for future research, and facilitate the design of more generalizable and reliable imitation learning systems.

\begin{acks}
The research described in this paper was carried out within the framework of the National Recovery and Resilience Plan Greece 2.0, funded by the European Union - NextGenerationEU (Implementation Body: HFRI. Project name: DEEP-REBAYES. HFRI Project Number 15430).
\end{acks}
%%
%% The next line prints the references.
\printbibliography

% \appendix

% \section{Reproducibility Checklist for JAIR}

% Select the answers that apply to your research -- one per item. 

% \subsection*{All articles:}

% %\hh{revised for stylistic consistency:}
% \begin{enumerate}
%     \item All claims investigated in this work are clearly stated. 
%     [\textbf{yes}]
%     \item Clear explanations are given how the work reported substantiates the claims. 
%     [\textbf{yes}]
%     \item Limitations or technical assumptions are stated clearly and explicitly. 
%     [\textbf{yes}]
%     \item Conceptual outlines and/or pseudo-code descriptions of the AI methods introduced in this work are provided, and important implementation details are discussed. 
%     [\textbf{yes}]
%     \item 
%     Motivation is provided for all design choices, including algorithms, implementation choices, parameters, data sets and experimental protocols beyond metrics.
%     [\textbf{yes}]
% \end{enumerate}

% \subsection*{Articles containing theoretical contributions:}
% Does this paper make theoretical contributions? 
% [\textbf{no}]

% \subsection*{Articles reporting on computational experiments:}
% Does this paper include computational experiments? [\textbf{no}]

% \subsection*{Articles using data sets:}
% Does this work rely on one or more data sets (possibly obtained from a benchmark generator or similar software artifact)? 
% [\textbf{no}]

\end{document}